\documentclass[sigconf]{acmart}
\AtBeginDocument{%
  \providecommand\BibTeX{{%
    Bib\TeX}}}

\usepackage{soul}
\usepackage{url}
\usepackage{graphicx}
\usepackage{amsmath}
\usepackage{amsthm}
\usepackage{booktabs}
\usepackage{algorithm}
\usepackage{algorithmic}
\usepackage[switch]{lineno}
\usepackage{makecell}
\usepackage{pdflscape}
\usepackage{forest}
\usepackage{balance}
\usepackage{flushend}
\usepackage{multicol}

\copyrightyear{2025}
\acmYear{2025}
\setcopyright{acmlicensed}\acmConference[KDD '25]{Proceedings of the 31st ACM SIGKDD Conference on Knowledge Discovery and Data Mining V.2}{August 3--7, 2025}{Toronto, ON, Canada}
\acmBooktitle{Proceedings of the 31st ACM SIGKDD Conference on Knowledge Discovery and Data Mining V.2 (KDD '25), August 3--7, 2025, Toronto, ON, Canada}
\acmDOI{10.1145/3711896.3736571}
\acmISBN{979-8-4007-1454-2/2025/08}

\begin{document}

\title{A Survey on Deep Learning based Time Series Analysis with Frequency Transformation}



\author{Kun Yi}
\email{kunyi.cn@gmail.com}
\affiliation{%
  \institution{State Information Center}
  \city{Beijing}
  \country{China}
}

\author{Qi Zhang*}
\thanks{*Corresponding Author}
\email{zhangqi_cs@tongji.edu.cn}
\affiliation{%
  \institution{Tongji University}
  \city{Shanghai}
  \country{China}}

\author{Wei Fan*}
\email{wei.fan@auckland.ac.nz}
\affiliation{%
  \institution{University of Auckland}
  \city{Auckland}
  \country{New Zealand}
}

\author{Longbing Cao}
\email{longbing.cao@mq.edu.au}
\affiliation{%
 \institution{Macquarie University}
 \city{Sydney}
 \country{Australia}}

\author{Shoujin	Wang}
\email{shoujin.wang@uts.edu.au}
\affiliation{%
  \institution{University of Technology Sydney}
  \city{Sydney}
  \country{Australia}}

\author{Hui	He}
\email{hehui617@bit.edu.cn}
\affiliation{%
  \institution{Beijing Institute of Technology}
  \city{Beijing}
  \country{China}}

\author{Guodong	Long}
\email{guodong.long@uts.edu.au}
\affiliation{%
  \institution{University of Technology Sydney}
  \city{Sydney}
  \country{Australia}}

\author{Liang Hu}
\email{milkrain@gmail.com}
\affiliation{%
  \institution{Tongji University}
  \city{Shanghai}
  \country{China}}

\author{Qingsong Wen}
\email{qingsongedu@gmail.com}
\affiliation{%
  \institution{Squirrel Ai Learning}
  \city{Bellevue}
  \country{USA}}
  
\author{Hui	Xiong}
\email{xionghui@ust.hk}
\affiliation{%
  \institution{Hong Kong University of Science and Technology (Guangzhou)}
  \city{Guangzhou}
  \country{China}}

\renewcommand{\shortauthors}{Kun Yi, et al.}

\begin{abstract}
Recently, frequency transformation (FT) has been increasingly incorporated into deep learning models to significantly enhance state-of-the-art accuracy and efficiency in time series analysis. The advantages of FT, such as high efficiency and a global view, have been rapidly explored and exploited in various time series tasks and applications, demonstrating the promising potential of FT as a new deep learning paradigm for time series analysis. Despite the growing attention and the proliferation of research in this emerging field, there is currently a lack of a systematic review and in-depth analysis of deep learning-based time series models with FT. It is also unclear why FT can enhance time series analysis and what its limitations are in the field. 
To address these gaps, we present a comprehensive review that systematically investigates and summarizes the recent research advancements in deep learning-based time series analysis with FT. Specifically, we explore the primary approaches used in current models that incorporate FT, the types of neural networks that leverage FT, and the representative FT-equipped models in deep time series analysis. We propose a novel taxonomy to categorize the existing methods in this field, providing a structured overview of the diverse approaches employed in incorporating FT into deep learning models for time series analysis. Finally, we highlight the advantages and limitations of FT for time series modeling and identify potential future research directions that can further contribute to the community of time series analysis.
\end{abstract}





\maketitle

\section{Introduction}
Time series data is amongst the most ubiquitous data types, and has penetrated nearly every corner of our daily life~\cite{FDama_forecast_2021}, e.g., user-item interaction series in e-commerce and stock price series over time in finance. In recent years, time series analysis has attracted rapidly increasing attention from academia and industry, particularly in areas such as time series forecasting~\cite{Benidis_forecasting_2022}, anomaly detection~\cite{Zahra_anomaly_2022}, and classification~\cite{FawazFWIM19}. Time series analysis has played a critical role in a wide variety of real-world applications to address significant challenges around us long-lastingly, such as traffic monitoring~\cite{Bai2020nips,deepcn_2024}, financial analysis~\cite{FengHWLLC19, Amplifier_2025}, and COVID-19 prediction~\cite{tampsgcnets2022,HeZYSNC25}. However, time series analysis is extremely challenging due to the intricate inter-series correlations and intra-series dependencies. 

\begin{figure}
    \centering
    \includegraphics[width=1\linewidth]{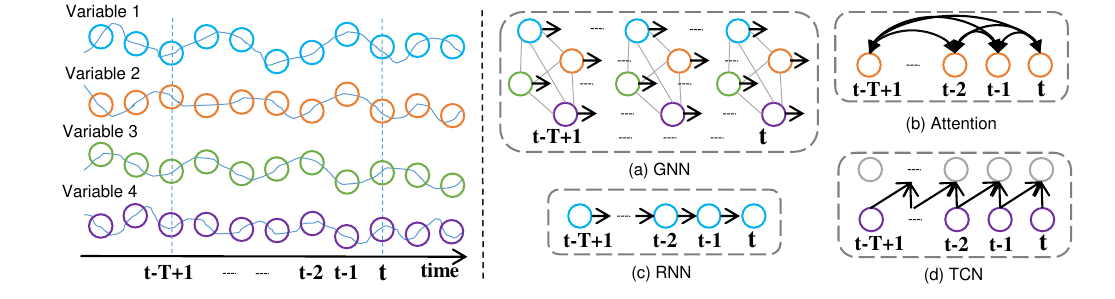}
    \caption{Illustration of various working mechanisms applied to time series data. We take an example of four variables and $T$ timestamps, as shown in the left portion of the figure. (a) GNN constructs a graph connecting variables for each timestamp. (b) Self-attention builds temporal connections for each variable. (c) RNN creates a recursive cycle for capturing temporal transitions. (d) TCN consists of a stack of causal convolutional layers over timestamps.}
    \label{fig:time_domain_problem}
    \vspace{-7mm}
\end{figure}

Previous time series models based on deep learning have been devoted to modeling complex intra- and inter-series dependencies in the time domain to enhance downstream tasks. 
Representative sequential models such as recurrent neural networks (RNNs)~\cite{LSTNet_2018,LSTMNAD_2018}, temporal convolutional networks (TCNs)~\cite{TCN_2018}, and attention networks~\cite{autoformer_2021} are utilized to capture intra-series dependencies, while convolutional networks such as convolutional neural networks (CNNs)~\cite{DCRNN_2018} and graph neural networks (GNNs)~\cite{Chen_2022} are preferred to attend to inter-series correlations. Although achieving good results, those networks have inherent drawbacks of time-domain modeling, limiting their capabilities in capturing critical patterns for time series analysis. For example, GNNs are constructed based on variable-wise connections as illustrated in Fig. \ref{fig:time_domain_problem}(a), and the sequential models (i.e., Transformer, RNN, and TCN) are based on timestamp-wise connections as shown in Fig. \ref{fig:time_domain_problem}(b), (c), and (d), respectively. These modelings consider point-wise (e.g., variable/timestamp-wise) connections and fail to attend to whole or sub time series. Therefore, they are usually incapable of modeling common but complex global patterns, such as periodic patterns of seasonality, in time series~\cite{ATFN_2022,CoST_2022}. These inherent drawbacks inspire researchers to address the intricate inter-series correlations and intra-series dependencies of time series from a different perspective.

Recently, deep learning methods leveraging frequency transformation (FT)~\cite{roberts1987digital}, e.g., Discrete Fourier Transform (DFT)~\cite{winograd1976computing}, Discrete Cosine Transform (DCT)~\cite{ahmed1974discrete}, and Discrete Wavelet Transform (DWT)~\cite{shensa1992discrete}, have gained a surge of interest within the machine learning community~\cite{XuCVPR2020,ChiJM20,John2021,ZhouYZW22}. These neural models incorporating frequency transformation have demonstrated an efficient learning paradigm in time series analysis and achieved state-of-the-art performance in terms of both efficiency and effectiveness~\cite{autoformer_2021,fedformer_2022,TFAD_2022}. This can be attributed to the distinctive advantages of FT (see Section \ref{ft_advantage}) that the frequency spectrums generated by FT contain abundant vital patterns, e.g., seasonal trends, and provide a global view of the characteristics of time series.
In addition, FT facilitates obtaining multi-scale representations and multi-frequency components of time series for capturing informative representations and patterns. This motivates us to systematically summarize and analyze the advantages of FT to instruct researchers in this area and to deliver a comprehensive survey on the emerging area, i.e., deep learning based time series analysis with FT, thereby enlightening the time series community. 
While the literature includes various studies that discuss time series analysis from different perspectives~\cite{Fakhrazari_2017,Benidis_forecasting_2022,FawazFWIM19,ChenPZS21,ChenKS22,0001EL21,liang_kdd2024,zhang_pami2024,Wang_arxiv_2024}, there remains a lack of comprehensive summaries on the topic of time series analysis with FT. 
Moreover, the reasons why FT can enhance the time series analysis have not yet been summarized, and its limitations have not been thoroughly analyzed. These gaps have hindered the theoretical development and practical applications of time series analysis with FT.

\begin{figure*}[!t]
	\centering
    \includegraphics[width=0.85\linewidth]{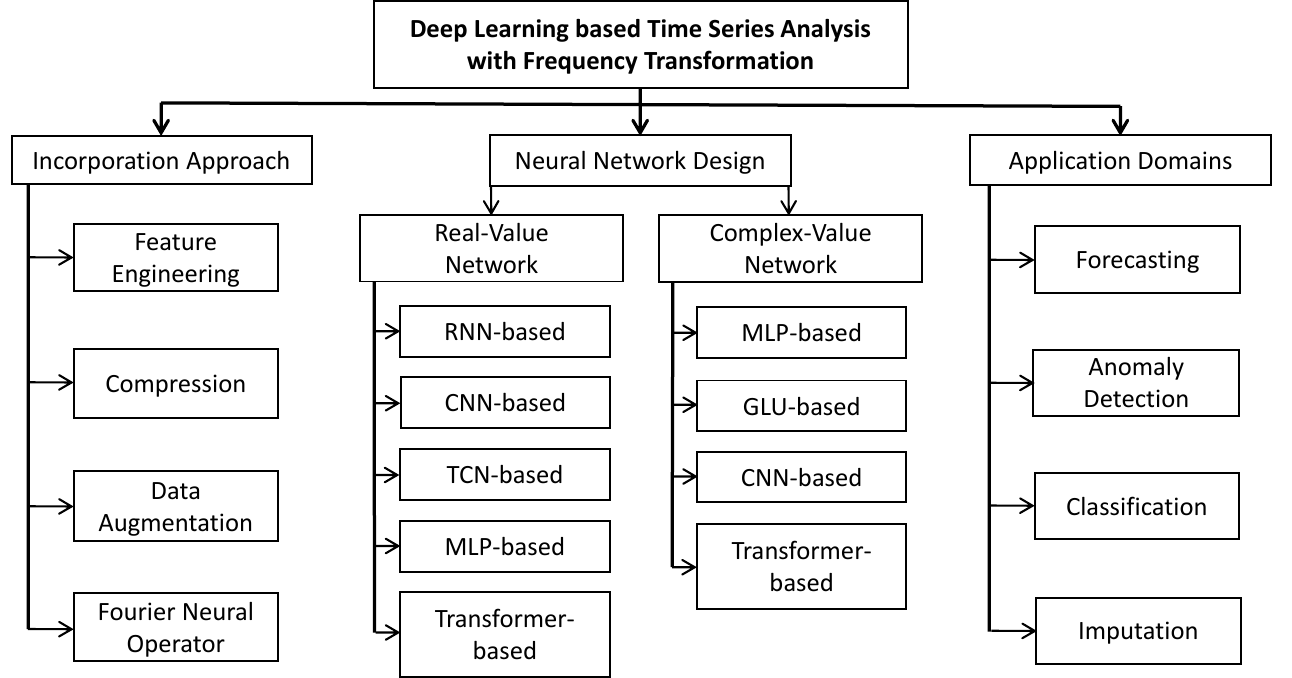}
	\caption{A taxonomy of deep learning based time series analysis with frequency transformation.  
	}
	\label{fig:taxonomy}
\end{figure*}

In this paper, we aim to fill the aforementioned gaps by reviewing existing deep learning methods for time series analysis with FT. Specifically, our primary objective is to provide answers to four crucial perspectives: i) the strategies employed by current neural time series models in incorporating neural networks with FT; ii) the specific types of neural networks utilized in conjunction with FT; iii) the representative FT-equipped neural models commonly employed in time series applications; and iv) an exploration of the reasons behind FT to enhance neural models as well as an analysis of its limitations in the context of time series analysis.
By addressing these questions, we provide valuable insights into the realm of neural time series analysis with FT. To the best of our knowledge, this paper is the first work to comprehensively and systematically review neural time series analysis with FT and to propose a new taxonomy for this emerging area, as depicted in Fig. \ref{fig:taxonomy}. 


\section{Preliminaries}\label{sec:background}
\subsection{Time Series Analysis}
In this section, we provide a brief introduction to the four fundamental tasks of time series analysis before diving into neural time series analysis with Fourier Transform (FT). 

\subsubsection{Forecasting}
Time series forecasting is the task of predicting future data points based on past observations~\cite{Benidis_forecasting_2022}. Consider a dataset with $N$ different time series measured over $T$ time steps, represented as $\mathbf{X} = [X_1, X_2, ..., X_T] \in \mathbb{R}^{N \times T}$, where each $X_t \in \mathbb{R}^N$ captures the values of all series at time $t$. To make predictions at time $t$, we use a window of past $L$ observations, $\mathbf{X}_t = [X_{t-L+1}, ..., X_t]$, as input. The goal is to forecast the next $\tau$ values, denoted $\mathbf{Y}_t = [X_{t+1}, ..., X_{t+\tau}]$. A forecasting model, $f_\theta$, learns to map $\mathbf{X}_t$ to an estimate $\hat{\mathbf{Y}}_t = f_\theta(\mathbf{X}_t)$.

\subsubsection{Classification}
Time series classification aims to assign a categorical label to each time series in a dataset~\cite{FawazFWIM19}. Formally, consider a dataset $D = \{(X_1, Y_1), (X_2, Y_2), \ldots, (X_N, Y_N)\}$, where each $X_i \in \mathbb{R}^T$ represents a univariate time series with $T$ time steps, and $Y_i$ is its associated one-hot encoded label. Assuming $D$ contains $K$ distinct classes, each label $Y_i \in \mathbb{R}^K$ is a binary vector such that the $j$-th element is 1 if $X_i$ belongs to class $j$, and 0 otherwise. The goal is to learn a classifier $f_\theta$, parameterized by $\theta$, that maps input time series to class probability distributions: $Y_i = f_\theta(X_i)$.

\subsubsection{Anomaly Detection}
Time series anomaly detection aims to identify abnormal patterns or subsequences within a temporal sequence~\cite{ChenPZS21}. The objective is to design models or algorithms capable of distinguishing between normal and anomalous behaviors, enabling early detection and timely alerts for unusual events. Given a time series $X = [x_1, x_2, \ldots, x_T]$ with $T$ timestamps, where $x_i$ denotes the observation at time $i$, the task is to identify a subset $X_s \subseteq X$ corresponding to anomalous data points that deviate significantly from expected temporal patterns.
\subsubsection{Imputation}
Time series imputation is the task of estimating missing values in a partially observed time series. Formally, let
\({X} = [\,{x}_1,\ldots,{x}_T]\in\mathbb{R}^T\)
denote the observed series (possibly with missing entries), and let
\(M \in \{0,1\}^T\)
be a binary mask such that \(m_t = 0\) indicates a missing value at index \(t\), and \(m_t = 1\) indicates an observed value. The goal is to recover a complete time series
\(\hat{X} = [\,\hat{x}_1,\ldots,\hat{x}_T]\in\mathbb{R}^T\)
that approximates the underlying true series while filling in the unobserved entries. A typical imputation model
\(f_\theta\) (parameterized by \(\theta\))
maps the partially observed data and mask to the reconstructed output,
$\hat{X} = f_\theta({X}, M), $
such that
\(\hat{x}_t \approx {x}_t\)
for all observed entries and
\(\hat{x}_t\)
is inferred for missing ones.

\subsection{Frequency Transformation}
In this section, we briefly introduce commonly used frequency transformation that converts time-domain data into the frequency domain, including Discrete Fourier Transform (DFT), Discrete Cosine Transform (DCT), and Discrete Wavelet Transform (DWT). Additionally, we describe the convolution theorem, which is a fundamental property in the frequency domain.

\subsubsection{Discrete Fourier Transform}
Discrete Fourier Transform (DFT) plays an important role in the area of digital signal processing. Given a sequence $x[n]$ with the length of N, DFT converts $x[n]$ into the  frequency domain by:
\begin{equation}\label{equ:dft}
    \mathcal{X}[k] = \sum_{n=0}^{N-1}x[n]e^{-j(2\pi/N)kn},\ s.t.,\ k=0,1,...,N-1
\end{equation}
where $j$ is the imaginary unit and $\mathcal{X}[k]$ represents the spectrum of $x[n]$ at the frequency $\omega_k=2\pi k/N$. The spectrum $\mathcal{X} \in \mathbb{C}^{k}$ consists of real parts $\operatorname{Re}=\sum_{n=0}^{N-1}x[n]\cos{(2\pi/N)kn} \in \mathbb{R}^{k}$ and imaginary parts $\operatorname{Im}= -\sum_{n=0}^{N-1}x[n]\sin{(2\pi/N)kn} \in \mathbb{R}^{k}$ as:
\begin{equation}\label{dft_r_i}
    \mathcal{X}=\operatorname{Re}+j\operatorname{Im}.
\end{equation}
The amplitude part $A$ and phase part $\theta$ of $\mathcal{X}$ is defined as:
\begin{equation} \label{equ_amplitude}
    A=\sqrt{\operatorname{Re}^2+\operatorname{Im}^2}.
\end{equation}
\begin{equation} \label{equ_phase}
    \theta=\arctan (\frac{\operatorname{Im}}{\operatorname{Re}}).
\end{equation}

\subsubsection{Discrete Cosine Transform}

Discrete Cosine Transform (DCT) has emerged as the de-facto image transformation in most visual systems. The most common 1-D DCT $C(k)$ of a data sequence $x[n]$ is defined as:
\begin{equation}\label{equ:dct}
    C(k)=\alpha(k)\sum_{n=0}^{N-1}x[n]\cos\left [ \frac{\pi(2n+1)k}{2N}  \right ]  
\end{equation} where $k=0,1,...,N-1$, and $\alpha(k)$ is defined as:
\begin{equation}
    \alpha (k)= \left\{\begin{matrix} 
   \sqrt{\frac{1}{N} },for\quad k=0 \\  
   \sqrt{\frac{2}{N} },for\quad k\ne0 
\end{matrix}\right.
\end{equation}
DCT only retains the real parts of DFT, and 
often performs on real data with even symmetry or in some variants where the input or output data are shifted by half a sample.

\subsubsection{Discrete Wavelet Transform}

Discrete Wavelet Transform (DWT) has been shown to be an appropriate tool for time-frequency analysis. It decomposes a given signal into a number of sets in which each set is a time series of coefficients describing the time evolution of the signal in the corresponding frequency band.

For a signal $x(t)$, the wavelet transform $\operatorname{WT}$ can be expressed as $\operatorname{WT}(a,b)=\int_{-\infty}^{\infty}x(t)\Psi_{a,b}(t)\mathrm{d}t  = \left \langle x(t) ,\Psi_{a,b}(t)\right \rangle$
where $\Psi$ is the wavelet basis function. The basis generation can be defined by $\Psi_{a,b}(t)=\frac{1}{\sqrt{a}}\Psi\left ( \frac{t-b}{a }  \right )$
where $a$ and $b$ are the scaling and translation factors respectively. $\operatorname{DWT}$ discretizes the scale factor a and the translation factor b as $a=a_0^m,b=ka_0^mb_0, m,k \in \mathbb{Z}$.
Typically, $a_0$ is set to 2, and $b_0$ is set to 1. Accordingly, the $\operatorname{DWT}$ can be defined as:
\begin{equation}\label{equ:dwt}
    \operatorname{DWT}(a,b)=a_0^{-m/2}\int_{-\infty}^{\infty}x(t)\Psi(a_0^{-m}t-kb_0)(t)\mathrm{d}t. 
\end{equation}

In contrast to $\operatorname{DFT}$ and $\operatorname{DCT}$, $\operatorname{DWT}$ has the ability to identify the locations containing observed frequency content, while the DFT and DCT can only extract pure frequencies from the signal. Hence, $\operatorname{DWT}$ can perform time-frequency analysis. In addition, $\operatorname{DWT}$ can obtain different resolution representations~\cite{Mallat_1989} by changing the scaling and translation factors. In Table \ref{tab:compare_ft}, we compare the three frequency transformation methods, including their pros and cons. 

\begin{table}[!t]
    \centering
    \caption{Comparison of DFT, DCT, and DWT for time series analysis.}
    \vspace{-1mm}
    \scalebox{0.52}{
    \begin{tabular}{c c c c c c}
    \toprule[1.5pt]
       FT & \makecell[c]{Basis\\ Function} &  \makecell[c]{Value Type} & Time-Frequency & Pros & Cons\\
       \midrule[1pt]
       DFT & Sine+Cosine & Complex & No & Shift-invariant & \makecell[c]{Leakage effect \\ Lack of time localization}\\
       \midrule
       DCT & Cosine & Real & No & Computationally efficient & \makecell[c]{No phase information \\ Lack of time localization} \\
       \midrule
       DWT & Wavelet & Real & Yes & \makecell[c]{Multi-resolution analysis\\Localization in time and frequency} & Computation complexity\\
       \bottomrule[1.5pt]
    \end{tabular}
    }
    \label{tab:compare_ft}
    \vspace{-2mm}
\end{table}

\subsubsection{Convolution Theorem}
The convolution theorem \cite{1990Continuous} states that the Fourier transform of a circular convolution of two signals equals the point-wise product of their Fourier transforms. Given a signal $x[n]$ and a filter $h[n]$, the convolution theorem can be defined as follows:
\begin{equation}\label{convolution_theorem}
    \mathcal{F}(x[n]*h[n])=\mathcal{F}(x)\mathcal{F}(h)
\end{equation}
where $x[n]*h[n]=\sum_{m=0}^{N-1} h[m]x[(n-m)_N]$, $(n-m)_N$ denotes $(n-m)$ modulo N, and $\mathcal{F}(x)$ and $\mathcal{F}(h)$ denote discrete Fourier transform of $x[n]$ and $h[n]$, respectively.


According to the Convolution Theorem, the point-wise multiplication of the frequency spectra of two sequences corresponds to their circular convolution in the time domain. This operation, which inherently spans the entire sequence, enables more effective capture of global patterns such as periodicity, while also reducing computational cost~\cite{AlaaCS21}.

\section{Incorporation Approach}\label{sec:usage}
In this section, we present a systematic summary and discussion of the research categorization and progress regarding incorporating the frequency transformation to enhance time series analysis.

\subsection{Feature Engineering}
Previous works employ frequency transformation (DFT, DCT, and DWT) as feature engineering tools to obtain frequency domain patterns. Basically, they utilize frequency transformation to capture three primary types of information: periodic patterns, multi-scale patterns, and global dependencies.

\paragraph{\textbf{Periodicity}} Compared to the time domain, the frequency domain can provide vital information for time series, such as periodic information. Prior models take advantage of frequency domain information for periodic analysis and use it as an important complement to the time domain information. ATFN~\cite{ATFN_2022} proposes a frequency-domain block to capture dynamic and complicated periodic patterns of time series data, and integrates deep learning networks with frequency patterns. TFAD~\cite{TFAD_2022} utilizes a frequency domain analysis branch to detect complex pattern anomalies, e.g., periodic anomalies. CoST~\cite{CoST_2022} learns the trend representations in the time domain, whereas the seasonal representations are learned by a Fourier layer in the frequency domain. FreDo~\cite{fredo2022} is a frequency domain-based neural network model that is built on top of the baseline model to enhance its performance. ETSformer~\cite{etsformer2022} utilized DFT to design a frequency attention mechanism to replace the self-attention mechanism to identify seasonal patterns. 
DeepTime~\cite{WooLSKH23} leverages a novel concatenated Fourier features module to efficiently learn high-frequency patterns in time series.

\paragraph{\textbf{Multi-Scale}} One big challenge for time series analysis is that there are intricate entangled temporal dynamics among time series data. To address this challenge, some methods try to solve it in terms of the frequency domain. They disentangle temporal patterns by decomposing time series data into different frequency components. SFM~\cite{SFM_2017} separates the memory states of RNN into different frequency states such that they can explicitly learn the dependencies of both the low- and high-frequency patterns. \cite{ZhangAQ_kdd_2017} explicitly decomposes trading patterns into various frequency components and each component models a particular frequency of latent trading pattern underlying the fluctuation of stock price. 
Recently, wavelet-based models have shown competitive performances since wavelet transform can retain both time and frequency information and obtain multi-resolution representations. 
\cite{WangWLW_kdd_2018} proposes a wavelet-based neural network structure for building frequency-aware deep learning models for time series analysis.
\cite{Wen_sigmod_2021} applies maximal overlap discrete wavelet transform to decouple time series into multiple levels of wavelet coefficients and then detect single periodicity at each level.
\cite{WangYJ023} devises a novel data-dependent wavelet attention mechanism for dynamic frequency analysis of non-stationary time series analysis. 
\cite{YangLWZPW23} proposes an end-to-end graph enhanced Wavelet learning framework for long sequence forecasting which utilizes DWT to represent MTS in the wavelet domain.

\vspace{-2mm}
\paragraph{\textbf{Global Dependencies}} Existing time domain methods construct their models based on point-wise connections (see Fig. \ref{fig:time_domain_problem}), which prevent them from capturing series-level patterns, such as overall characteristics of time series. By leveraging the global view property of the frequency domain, some works utilize frequency information to attend to series-level patterns.
FEDformer~\cite{fedformer_2022} combines Fourier analysis with the Transformer which helps the Transformer better capture the global properties of time series.
TFAD~\cite{TFAD_2022} integrates the frequency domain analysis branch with the time domain analysis branch and detects seasonality anomalies in the frequency domain. Besides, some works introduce frequency domain analysis to improve neural networks in order to address their inherent drawbacks. 
Vanilla convolutions in modern deep networks are known to operate locally, which causes low efficacy in connecting two distant locations in the network. To mitigate the locality limitation of convolutions, 
SRL~\cite{Chi_2019} converts data into the frequency domain and proposes spectral residual learning for achieving a fully global receptive field, and FFC~\cite{ChiJM20} harnesses the Fourier spectral theory and designs an operation unit to leverage frequency information for enlarging the receptive field of vanilla convolutions. 

\subsection{Compression}  
Previous works utilize frequency transformation to obtain sparse representations and remove redundant information in the frequency domain. Moreover, since noise signals usually appear as high frequencies, it is easy to filter them out in the frequency domain.
For example, in FiLM~\cite{FiLM_2022}, authors view time series forecasting from the sequence compression perspective and apply Fourier analysis to keep the part of the representation related to low-frequency Fourier components to remove the impact of noises.
\cite{Rippel_nips_2015} proposes spectral pooling that performs dimensionality reduction by truncating the representation in the frequency domain because energy is heavily concentrated in the lower frequencies. 
\cite{XuCVPR2020} proposes a learning-based frequency selection method to identify the trivial frequency components while removing redundant information.

\vspace{-5mm}
\subsection{Data Augmentation}
Recently, a few studies have investigated data augmentation from a frequency domain perspective for time series~\cite{Wen0Y_augmentaion_2021}.
Since the frequency domain contains some vital information for time series analysis, such as periodic patterns, existing methods incorporate frequency domain features with time domain features for data augmentations with the aim of enhancing time series representations. 
For example, 
CoST~\cite{CoST_2022} incorporates a novel frequency domain contrastive loss which encourages discriminative seasonal representations and sidesteps the issue of determining the period of seasonal patterns present in the time series data.
BTSF~\cite{Yang_ICML_2022} fuses the temporal and spectral features to enhance the discriminativity and expressiveness of the representations.
TS-TFC~\cite{LiuMMW23} proposes a temporal-frequency co-training model for time-series semi-supervised learning, utilizing the complementary information from two distinct views for unlabeled data learning. 

More recently, different from CoST and BTSF that apply DFT after augmenting samples in the time domain, one new approach named TF-C~\cite{zhang_nips_2022} introduces frequency domain augmentations that directly perturb the frequency spectrum. 
It develops frequency-based contrastive augmentation to leverage rich spectral information and directly perturbs the frequency spectrum to leverage frequency-invariance for contrastive learning. Compared to performing data augmentations directly in the frequency domain (e.g., TF-C), applying the FFT after augmenting samples in the time domain (e.g., CoST and BTSF) may lead to information loss.
\vspace{-5mm}
\subsection{Fourier Neural Operator Learning}
According to the convolution theorem, differentiation is equivalent to multiplication in the Fourier domain~\cite{LiKALBSA2021}. This efficiency property makes DFT frequently used to solve differential equations.

Recently, Fourier Neural Operators (FNOs)~\cite{LiKALBSA2021}, which is currently the most promising one of the neural operators~\cite{Kovachki2021}, have been proposed as an effective framework to solve partial differential equations (PDEs).
More recently, FNO has been introduced in time series forecasting.
FEDformer~\cite{fedformer_2022} proposes Fourier-enhanced blocks and Wavelet-enhanced blocks to capture important structures in time series through frequency domain mapping.
FourierGNN~\cite{EVFGN_2022} reformulates the graph convolution operator in the frequency domain and efficiently computes graph convolutions over a hypervariate graph which represents the high-resolution correlations between any two variables at any two timestamps.

\section{Neural Network Design}\label{sec:network_design}
In this section, we delve deeper into existing related models that utilize specific types of neural networks to leverage frequency information. Considering that frequency transformation outputs can be either complex values or real values (as shown in Table \ref{tab:compare_ft}), and each value type requires distinct handling methods, we discuss the models from the perspectives of these two value types.
\vspace{-2.5mm}
\subsection{Complex-Value Data}
The DFT output values are complex and can be represented in two ways. One representation is through the real and imaginary parts (as shown in Equation (\ref{dft_r_i})), while the other representation is through the amplitude and phase parts (as depicted in Equations (\ref{equ_amplitude}) and (\ref{equ_phase})). While it is possible to simplify the calculation by retaining only one part, such as discarding the imaginary components~\cite{GodfreyG_tnnls_2018}, this approach may result in information loss.

In fact, there are two main approaches for performing neural networks on complex values. One approach is to treat each part of the complex value as a feature and then feed them to neural networks, respectively. Afterward, the output of corresponding networks is combined as a complex type (e.g., like Equation (\ref{dft_r_i})), and then the inverse DFT is executed and transmitted to the time domain.
For example, StemGNN~\cite{StemGNN_2020} conducts GLU~\cite{glu2017} on real and imaginary parts, respectively, which concatenates them as a complex value and applies IDFT.
ATFN~\cite{ATFN_2022} utilizes two linear layers to process the amplitude and phase parts, respectively, and then combines them as a whole.
The other is to conduct complex multiplication in the frequency domain directly.
FEDformer~\cite{fedformer_2022} randomly samples a few frequencies and conducts complex multiplication with a parameterized kernel incorporated with attention architecture.

\vspace{-2.5mm}
\subsection{Real-Value Data}
The output value type of DCT and DWT is real, hence commonly used network structures can be directly applied to them. For instance, DEPTS~\cite{fanwei2022} performs DCT and can be seamlessly integrated with an MLP to process the output parameters of DCT. Although the output value type of DFT is complex, some work discards one part, such as the phase part~\cite{ZhangAQ_kdd_2017}, making their network design effectively real-valued. Additionally, some methods just apply filtering in the frequency domain and transform back to the time domain, resulting in network designs that also belong to real-valued networks.

Except for capturing frequency patterns, in contrast to other network designs, one main purpose of network design for frequency-based models is the frequency component selection to decide which component is discriminative or critical. 
For example, 
RobustPeriod~\cite{Wen_sigmod_2021} applies DWT to decouple time series into multiple levels of wavelet coefficients and then proposes a method to robustly calculate unbiased wavelet variance at each level and rank periodic possibilities. TimesNet~\cite{timesnet_2023} selects the top-$k$ amplitude values while discarding the remaining ones. FAN~\cite{fan_2024} decomposes frequency components into non-stationary and stationary parts and employs specialized network modules for each.

\vspace{-2mm}
\section{Applications}\label{sec:application}
In this section, we review the representative FT-equipped neural time series models. We categorize them into four main applications: forecasting, anomaly detection, classification, and imputation. In Table \ref{tab:compare_modes}, we further compare them from six dimensions.

\begin{table*}[!t]
    \centering
    \renewcommand\arraystretch{1.5}
    \caption{Summary of representative FT-equipped deep learning models in time series analysis.}
    \scalebox{0.95}{
    \begin{tabular}{l c c c c c c}
    \toprule[1.5pt]
       Models  &  \makecell[c]{Frequency \\ Transformation}  & \makecell[c]{Incorporation\\ Approach} & Value Type & \makecell[c]{Neural\\ Network} & \makecell[c]{Application \\Domains} & \makecell[c]{Leveraged\\ Advantages}\\
      \midrule[1pt]
       SFM~\cite{ZhangAQ_kdd_2017} & DFT & Feature engineering & Real-value & RNN & Forecasting  & Decomposition\\
       StemGNN~\cite{StemGNN_2020}  & DFT  & Feature engineering & Complex-value & GLU & Forecasting & Decomposition\\
       Autoformer~\cite{autoformer_2021}  & DFT & Feature engineering & Complex-value & Transformer & Forecasting & \makecell[c]{Global view\\Efficiency} \\
       DEPTS~\cite{fanwei2022} & DCT & Feature engineering & Real-value & MLP& Forecasting & Decomposition\\
       FEDformer~\cite{fedformer_2022}  & DFT  & \makecell[c]{Feature engineering \\Operator learning}  & Complex-value & Transformer &Forecasting & \makecell[c]{Global view\\Efficiency}\\
       CoST~\cite{CoST_2022} & DFT & Data augmentation & Complex-value & MLP &Forecasting & Decomposition\\
       FiLM~\cite{FiLM_2022} & DFT & Compression & Complex-value & MLP& Forecasting & \makecell[c]{Sparse \\Representation}\\ 
       WAVEFORM~\cite{YangLWZPW23} & DWT & Feature engineering & Real-value & GCN & Forecasting & Decompostion\\
       FourierGNN~\cite{EVFGN_2022} & DFT  & Operator learning & Complex-value & MLP & Forecasting & Efficiency\\
       FreTS~\cite{frets_2023} & DFT& Operator learning & Complex-value& MLP& Forecasting&Global view\\
       FilterNet~\cite{filternet_2024} & DFT& Feature engineering& Complex-value& MLP& Forecasting& Decomposition\\
       FreDF~\cite{fredf_2025} & DFT& Feature engineering& Complex-value& MLP& Forecasting& Decomposition\\
       \midrule[1pt]
       SR-CNN~\cite{Ren_KDD_2019} & DFT & Feature engineering & Real-value & CNN & \makecell[c]{Anomaly \\Detection} & Decomposition\\
       RobustTAD~\cite{Gao_2020} & DFT & Data augmentation & Complex-value & CNN & \makecell[c]{Anomaly \\Detection} & Decomposition\\
       TFAD~\cite{TFAD_2022} & DWT & Feature engineering & Real-value & TCN & \makecell[c]{Anomaly \\Detection} & Decomposition\\
       FITS~\cite{fits_2024} & DFT& Compression & Complex-value& MLP&\makecell[c]{Forecasting \\ Anomaly Detection} & \makecell[c]{Sparse \\Representation} \\
       \midrule[1pt]
       RCF~\cite{WangWLW_kdd_2018}& DWT & Feature engineering & Real-value & CNN & Classification & Decomposition\\
       WD~\cite{Khan_nips_2018}& DWT & Feature engineering & Real-value & CNN & Classification & Decomposition\\
       BTSF~\cite{Yang_ICML_2022} & DFT & Data augmentation & Real-value & CNN&  \makecell[c]{Classification \\ Forecasting} & Decomposition\\
       TF-C~\cite{zhang_nips_2022} & DFT & Data augmentation & Real-value & Transformer&  Classification  & Decomposition\\
       TimesNet~\cite{timesnet_2023} & DFT  & Feature engineering & Real-value & CNN &  \makecell[c]{Classification \\ Forecasting \\ Anomaly Detection \\ Imputation} & Decompostion\\
       \midrule
       FGTI~\cite{fgti_2024} & DFT  & Feature engineering & Complex-value & Transformer &  Imputation & Decompostion\\
       PSW-I~\cite{wang2025optimal} & DFT  & Feature engineering & Complex-value & MLP &  Imputation & Decompostion\\
    \bottomrule[1.5pt]
    \end{tabular}
    }
    \label{tab:compare_modes}
    \vspace{-1mm}
\end{table*}

\vspace{-3mm}
\subsection{Time Series Forecasting}
Time series forecasting is essential in various domains, such as decision making and financial analysis. 
Recently, some methods leverage frequency information to improve the accuracy or efficiency of time series forecasting.
SFM~\cite{ZhangAQ_kdd_2017} decomposes the hidden states of memory cells into multiple frequency components and models multi-frequency trading patterns.
StemGNN~\cite{StemGNN_2020} learns spectral representations which are easier to recognize after DFT.
Autoformer~\cite{autoformer_2021} leverages FFT to calculate auto-correlation efficiently.
DEPTS~\cite{fanwei2022} conducts DCT to extract periodic features and then applies multi-layer perceptrons to these features for periodicity dependencies in time series. 
FEDformer~\cite{fedformer_2022} captures the global view of time series in the frequency domain.
CoST~\cite{CoST_2022} learns the seasonal representations in the frequency domain.
FiLM~\cite{FiLM_2022} utilizes Fourier analysis to keep low-frequency Fourier components.
\textsc{DeRiTS}~\cite{wei_derits_2024} and FAN~\cite{fan_2024} address the non-stationary issues for time series forecasting from the frequency spectrum perspective.
FreDF~\cite{fredf_2025} mitigates label correlation by learning to forecast in the frequency domain, thereby reducing estimation bias.
\vspace{-2mm}
\subsection{Time Series Anomaly Detection}
Recently, frequency-based models have been introduced in anomaly detection.
RobustTAD~\cite{Gao_2020} explores the data augmentation methods in the frequency domain to further increase labeled data.
TFAD~\cite{TFAD_2022} takes advantage of frequency domain analysis for seasonality anomaly.
MACE~\cite{mace_2024} introduces a pattern extraction mechanism that exploits the inherent sparsity of the frequency domain to improve the model’s generalization across diverse normal patterns.
CATCH~\cite{wu2025catch} patchifys the frequency domain into distinct frequency bands, and adaptively discover channel correlation across frequency bands to effectively detect both point and subsequence anomalies.
FCVAE~\cite{FCVAE_2024} integrates global and local frequency features to capture long-periodic heterogeneous patterns and fine-grained short-periodic trends, enabling effective unsupervised anomaly detection in univariate time series.
TSAD~\cite{TSAD_2024} employs nested sliding windows, where the outer and inner windows correspond to the time and frequency domains to bridge the discrepancy between time and frequency representations.
\vspace{-2mm}
\subsection{Time Series Classification}
Time series classification is an important and challenging problem in time series analysis. Recently, a few models have considered frequency domain information to perform this task. 
RCF~\cite{WangWLW_kdd_2018} extracts distinguishing features from the DWT decomposed results.
WD~\cite{Khan_nips_2018} uses wavelet functions with adjustable scale parameters to learn the spectral decomposition directly from the signal.
BTSF~\cite{Yang_ICML_2022} fuses time and spectral information to enhance the discriminativity and expressiveness of the representations.
TF-C~\cite{zhang_nips_2022} develops frequency-based contrastive augmentation to leverage rich spectral information and explore time-frequency consistency in time series.
TSLANet~\cite{TSLANet_2024} leverages the power of Fourier transform alongside global and local filters to cover the whole frequency spectrum, while adaptively removing high frequencies that tend to introduce noises. 

\vspace{-2mm}
\subsection{Time Series Imputation}
Time series imputation refers to the process of filling in missing values in a time-dependent dataset. FGTI~\cite{fgti_2024} employs a frequency-aware diffusion model that uses high-frequency filters for residual imputation and dominant-frequency filters for trend and seasonal refinement. PSW-I~\cite{wang2025optimal} This work presents a novel Proximal Spectrum Wasserstein (PSW) discrepancy, which integrates pairwise spectral distance and selective matching regularization to accurately quantify distributional discrepancies between two sets of time series, specifically tailored for the imputation task. Both methods incorporate frequency technology to enhance the accuracy of imputations by addressing different components of the time series.

\section{Summary of Frequency Transformation}\label{sec:characteristics}

In this section, to investigate why FT can enhance the neural models and its limitations for time series analysis, 
we summarize the advantages and limitations of frequency transformation. 
\subsection{Advantages}\label{ft_advantage}
\paragraph{\textbf{Decomposition}} Frequency transformation can decompose the original time series into different frequency components that embody vital time series information, such as periodic patterns of seasonality. In particular, DWT can decompose a time series into a group of sub-series with frequencies ranked from high to low and obtain multi-scale representations. 
By decomposing time series in the time domain into different components in the frequency domain, it is naturally helpful to figure out and obtain beneficial information for time series analysis.

\paragraph{\textbf{Global View}} According to Equations (\ref{equ:dft}), (\ref{equ:dct}), and (\ref{equ:dwt}), a frequency spectrum is calculated through the summation of all signals over time. Accordingly, each spectrum element in the frequency domain attends to all timestamps in the time domain, illustrating that a spectrum has a global view of the whole sequence of time series. In addition, according to the convolution theorem (see Equation (\ref{convolution_theorem})), the point-wise product of frequency spectrums also captures the global characteristics of the whole sequence, inspiring to parameterize global learnable filters in the frequency domain.

\paragraph{\textbf{Sparse Representation}} Frequency transformation enables the provision of sparse representations for sequences. Taking DFT as an example, a substantial number of coefficients are close to zero, indicating that we can employ a reduced number of coefficients to represent the entire sequence. In other words, the corresponding representations in the frequency domain have a property of \textit{energy compaction}. For example, the important features of signals captured by a subset of DWT coefficients are typically much smaller than the original. Specifically, using DWT, it ends up with the same number of coefficients as the original signal where many of the coefficients may be close to zero. As a result, we can effectively represent the original signal using only a small number of non-zero coefficients.

\paragraph{\textbf{Efficiency}}
As mentioned earlier, frequency transformation often leads to sparse representations, where a substantial number of coefficients are close to zero. Exploiting this sparsity allows for efficient computations by discarding or compressing the negligible coefficients, resulting in reduced memory requirements and faster processing.
Moreover, according to the convolution theorem, convolution in the time domain corresponds to Hadamard's point-wise product in the frequency domain, which allows for convolution to be calculated more efficiently in the frequency domain. 
Therefore, considering the equivalence of the convolution theorem, convolution calculated in the frequency domain involves significantly fewer computational operations.

\subsection{Limitations}
\paragraph{\textbf{Loss of temporal information}} Frequency transformation techniques, including DFT and DCT, primarily emphasize capturing the frequency characteristics of a time series. While these techniques offer valuable insights into the frequency domain, they may overlook or inadequately represent temporal information~\cite{yang2024rethinking}. Certain temporal patterns or dynamics inherent in time series may not be fully captured in the frequency domain, thereby limiting the comprehensive analysis and understanding of temporal aspects~\cite{GodfreyG_tnnls_2018}.

\paragraph{\textbf{Dependence on pre-defined parameters}} Frequency transformation techniques often require setting parameters, such as window size, sampling rate, or frequency bands. Selecting appropriate parameter values can be challenging, and suboptimal choices may lead to inaccurate frequency representations or missing important frequency components~\cite{Khan_nips_2018,pnas_2022}. Accordingly, parameter tuning and optimization are necessary to ensure the effectiveness of frequency transformation in time series analysis.

\vspace{-2mm}
\section{Discussion for Future Opportunities}\label{sec:future}
In this section, we explore the prospects for future research in neural time series analysis with frequency transformation. We begin by outlining the current limitations of frequency transformation and propose innovative directions to overcome these challenges. Subsequently, we delve into open research issues and emerging trends in the field of time series analysis that can be addressed through the utilization of frequency transformations.
\vspace{-2mm}
\subsection{The Perspective of Frequency Transformation}
\subsubsection{Leveraging New Orthogonal Transform Technology}
Recent studies have shown the efficiency and effectiveness of orthogonal transform which serves as a plug-in operation in neural networks, including frequency analysis and polynomial family. Some new orthogonal transform technologies have been introduced in neural networks and achieved good results. For example, 
\cite{Park_kdd_2021,park_2024} propose the Partial Fourier Transform (PFT), an efficient and accurate algorithm designed to compute only a subset of Fourier coefficients, rather than the full spectrum.
The Fractional Fourier transform (FrFT) has been proven to be desirable for noise removal and can enhance the discrimination between anomalies and background~\cite{TaoZLLD_2019}.
In \cite{ZhaoTLPL_2022}, the authors utilize FrFT to enhance efficient feature fusion and comprehensive feature extraction.
\cite{zhao2022} leverages FrFT to enable flexible extraction of global contexts and sequential spectral information.
In the future, it would be a promising direction to incorporate more new orthogonal transform technologies for deep learning in time series analysis, such as FrFT.
\vspace{-1mm}
\subsubsection{Integrating Frequency Transformation with Deep Learning}
The basis functions used in frequency transformation, such as sine, cosine, and wavelet functions, are fixed across different domains. As a result, the frequency features extracted through these basis functions are domain-invariant.
In other words, the features are insensitive to unexpected noise or to changing conditions.
Few previous works combine frequency transformation with the learning ability of neural networks. 
\cite{Khan_nips_2018} proposes a method to efficiently optimize the parameters of the spectral decomposition based on the wavelet transform in a neural network framework. \cite{pnas_2022} mimics the fast DWT cascade architecture utilizing the deep learning framework. These methods have shown promising performances, and in the future, the combination of frequency transformation with deep learning deserves further investigation.
\vspace{-1mm}
\subsubsection{Jointly Learning in the Time and Frequency Domain}
The frequency domain only uses periodic components, and thus cannot accurately model the non-periodic aspects of a signal, such as a linear trend~\cite{GodfreyG_tnnls_2018}. Moreover, according to the uncertainty principle~\cite{TFAD_2022}, designing a model with a single structure that can capture the time and frequency patterns simultaneously is difficult.
As a result, in the future, an interesting direction is to take advantage of corresponding characteristics of learning in the time and frequency domain to improve the accuracy and efficiency of time series analysis. Few works have tried to learn representations in the time and frequency domain, respectively. 
More time-frequency representation learning methods are required in the future.
\vspace{-1mm}
\subsection{The Perspective of Time Series Analysis}
\subsubsection{Applying Frequency Transformation to Enhance Time Series Applications} 
Applying frequency transformation techniques to a broader range of time series applications has the potential to unlock valuable insights and enhance decision-making in various domains. No matter in detecting anomalies in physiological signals, uncovering market cycles in financial data, or identifying patterns in environmental parameters, frequency transformation enables a deeper understanding of complex temporal patterns and trends. By harnessing the power of frequency analysis, researchers and practitioners can uncover hidden relationships, improve forecasting accuracy, optimize resource management, and advance knowledge in diverse fields, ultimately driving innovation and enabling data-driven decision-making in a wide range of time series applications.
\vspace{-3mm}
\subsubsection{Scalability} Scalability~\cite{KeoghK03} is a key consideration in time series analysis. When coupled with frequency transformation techniques, it offers the potential for efficient and scalable analysis of large-scale time series data. Frequency transformation allows for the extraction of frequency components, reducing the dimensionality of the data and enabling more efficient processing. This dimensionality reduction can significantly improve the scalability of time series analysis algorithms, as it reduces computational complexity and memory requirements. 
Scalable time series analysis with frequency transformation can pave the way for analyzing and extracting insights from big data time series applications in domains such as the Internet of Things (IoT), financial markets, or sensor networks.

\vspace{-1mm}
\subsubsection{Privacy-preserving} Leveraging frequency transformation offers a powerful approach to data privacy-preserving~\cite{DworkMNS16} in time series analysis. By applying frequency transformation, time series data can be transformed into frequency domain representations without revealing the underlying raw data. This transformation allows for the extraction of frequency components and patterns while maintaining the confidentiality of the original information. Privacy-preserving with frequency transformation techniques can ensure individual privacy and data confidentiality, and enable collaborative analysis, data sharing, and research collaborations while mitigating privacy risks. This approach is particularly valuable in domains where data sensitivity is critical, such as healthcare, finance, or personal monitoring, allowing for the utilization of frequency analysis while protecting the privacy of individuals or organizations.

\section{Conclusion}
In this paper, we present a comprehensive survey on deep learning based time series analysis methods that leverage frequency transformation. 
We organize the reviewed methods from the perspectives of incorporation approaches, neural network design, and application domains, and we summarize the advantages and limitations of frequency transformation for time series analysis.
To the best of our knowledge, this is the first systematic and in-depth review focused specifically on neural time series analysis with frequency transformation, aiming to provide valuable insights for the research community. To support further exploration, we also provide a curated repository of related resources, available at \url{https://github.com/aikunyi/time_series_frequency}.

\newpage
\begin{acks}
This project is funded by the State Grid Corporation of China headquarters science and technology project "Research on the optimization technology and platform of the company's natural monopoly business and competitive business operation based on social information" (Project No. 52180025000C-194-ZN).
\end{acks}



\bibliographystyle{ACM-Reference-Format}
\bibliography{ref}

\end{document}